\documentclass{sig-alt-release2}

\usepackage{lscape,url}
\usepackage{multirow,datetime}

\begin{document}
%
\title{A Cross-cultural Corpus of Annotated Verbal and\\ 
Nonverbal Behaviors in Receptionist Encounters}

\numberofauthors{3} 
%
\author{
%
%
\alignauthor
Maxim Makatchev\\
       \affaddr{Robotics Institute}\\
       \affaddr{Carnegie Mellon University}\\
       \affaddr{Pittsburgh, PA, USA}\\
       \email{mmakatch@cs.cmu.edu}
\alignauthor
Reid Simmons\\
       \affaddr{Robotics Institute}\\
       \affaddr{Carnegie Mellon University}\\
       \affaddr{Pittsburgh, PA, USA}\\
       \email{reids@cs.cmu.edu}
\alignauthor
Majd Sakr\\
       \affaddr{Carnegie Mellon University in Qatar}\\
       \affaddr{Doha, Qatar}\\
       \email{msakr@qatar.cmu.edu}
}

\maketitle

\begin{abstract}
We present the first annotated corpus of nonverbal behaviors in receptionist interactions, and the first nonverbal corpus (excluding the original video and audio data) of service encounters freely available online. Native speakers of American English and Arabic participated in a naturalistic role play at reception desks of university buildings in Doha, Qatar and Pittsburgh, USA. Their manually annotated nonverbal behaviors include gaze direction, hand and head gestures, torso positions, and facial expressions. We discuss possible uses of the corpus and envision it to become a useful tool for the human-robot interaction community.
\end{abstract}

\section{Introduction}

Behavioral realism has been one of the promising directions in the development of on-screen conversational agents and robots capable of natural language dialogue (see \cite{RichSidner2009} for an overview). For example, interactions with a robot receptionist that evoke user's social response are associated with better engagement and lower rate of breakdowns during information-seeking dialogues~\cite{Simmons2011}. A necessary step in designing such interactions is to identify behaviors with a potential to evoke a desired user response.

Data sources that can be used to harvest behavior candidates include ethnographic and controlled studies. Ethnographic studies provide an opportunity for collection of naturalistic conversational data, but often face the issues of unclear sample population and coarse granularity of captured data~\cite{BeebeCummings1996}. On the other hand, collecting high resolution data in a controlled setting may hamper spontaneity  and naturalness of the interaction. In general, data collection methodology can influence both the sociopragmatic choices, namely, what speech act to say, and their pragmalinguistic realization, namely, how to say it (see~\cite{BeebeCummings1996} for a discussion).

These methodological difficulties, combined with the challenges of annotating multimodal data, result in the lack of annotated corpora of naturalistic interactions  for many scenarios that are currently relevant for human-robot interaction research. The corpus of role plays between a visitor and a receptionist in a realistic environment that we present in this paper attempts to help fill this gap. 

In the next section, we describe related work on corpora of service encounters. After that, we introduce our data collection methodology and the annotation scheme we use. We conclude with the discussion of possible uses of the corpus.

\section{Corpora of service encounters}

Audio corpora of human service encounters have been used for analysis of linguistic and paralinguistic features, such as timing and prosody. For example, Vienna-Oxford International Corpus of English~(VOICE) \cite{voice} includes service encounters between speakers of English as a lingua franca. Audio recordings of Syrian shopping interactions were collected and analyzed by Traverso~\cite{Traverso2001}. Service encounters gathered in public offices and shops of Catalonia were examined with respect to how bilinguals negotiate code (language) of their interaction. Audio recordings have been used to analyze politeness strategies in shopping interactions (see, for example, \cite{Kong2010}). 

The importance of gaze (see~\cite{Montague2011} for an overview) and smile (see, for example, \cite{Kim2009}) in defining the outcome of the service interactions suggest the need for capturing and studying nonverbal behaviors in videos.
For instance, customers reported higher satisfaction when they interacted face-to-face with a bank teller who responded with contingent smile, rather than constant neutral or constant smiling expression~\cite{Kim2009}. The same data showed that amused and polite smiles differ with respect to their temporal properties~\cite{Hoque2011}. Analysis of verbal and nonverbal expressions in the videos of interethnic encounters of Korean retailers with Korean and African-American customers showed that these language communities had different perception of function of socially minimal and socially expanded encounters~\cite{Bailey1997}.  

Receptionist interactions, a subtype of service encounters, were analyzed with respect to their verbal content via role plays in~\cite{Chee2010}. Hewitt et al.~\cite{Hewitt2009} conducted discourse analysis of dialogues involving hospital receptionists. The openly accessible CUBE-G corpus of nonverbal behaviors from role plays of German and Japanese participants covers scenarios that may be relevant for service encounters, including first meeting, negotiation and status difference~\cite{Rehm2009}. The original Map Task~\cite{Anderson1991} and followup projects collect direction-giving dialogues that may be relevant to some receptionist encounters.

We were not able to find any nonverbal corpora of human receptionist interactions. With respect to availability, among all the corpora mentioned above only VOICE, CUBE-G and Map Task related corpora are freely accessible. Hence, our corpus may be the first annotated corpus of nonverbal behaviors in receptionist interactions, and the first nonverbal corpus (excluding the original video and audio data) of service encounters freely available online~\cite{ReceptionistCorpus2012}.


\section{Data collection}

\subsection{Participants}
We recruited via emails and posters in Education City, in Doha, Qatar and via announcements posted on bulletin boards across CMU campus in Pittsburgh, USA. The recruitment materials specified that we were looking for native speakers of American English or Arabic.  Majority of the participants (17 of 22) were university students, staff, or faculty. The participants filled demographic surveys and evaluated themselves on ten-item personality inventory (TIPI)~\cite{Gosling2003} and 20-item positive and negative affect scale (PANAS)~\cite{Watson1988}. The distribution of participants is shown in Table~\ref{table:participants}. 

\begin{table}[htb]
\begin{center}
\begin{tabular}{|l|l|l|l|}
\hline
\multirow{4}{*}{Doha} & \multirow{2}{*}{Arabic} & Females & 2\\
 & & Males & 6 \\ \cline{2-4}
 & \multirow{2}{*}{American English} & Females & 2\\
 & & Males & 3 \\ \hline
\multirow{4}{*}{Pittsburgh} & \multirow{2}{*}{Arabic} & Females & 1 \\
 & & Males & 1\\ \cline{2-4}
 & \multirow{2}{*}{American English} & Females & 5\\
 & & Males & 1 \\ \hline
\end{tabular}
\end{center}
\caption{Distribution of participants between Doha and Pittsburgh experiment sites}
\label{table:participants}
\end{table}

People apply different criteria when they report their native language and mother tongue~\cite{Laitin2000}. To control for this, we asked the participants to list the countries they lived in for more than a year, and their age at the time of moving in and out of the country. All but 3 participants (who were all in the American English condition in Doha) spent the majority of their lives in the country where their native language is a primary spoken language. A female participant in Doha changed her reported native language from American English to Tulu, after asking the experimenter a clarification question. Her data remains in the corpus although she is not included in the Table~\ref{table:participants}.

Mean age of participants in Doha was 25 years ($SD=7.8$). In Pittsburgh, average age was 28.7 years ($SD=12.7$). Native speakers of Arabic were on average 23.2 years old ($SD=4.2$), while average age of native speakers of American English was 30.9 years ($SD=12.5$).


\subsection{Procedure}

After filling out the questionnaires, one of the participants was asked to play the role of a receptionist while another was asked to imagine themselves as a first-time visitor looking for a particular location inside the building. The location was picked by the experimenter from the following list: library, restroom, cafeteria, student recreation room, a professor A's office, etc. Visitors were asked to seek help of the receptionist for directions using English and then to proceed towards their destination.

Most of the participant pairs were not familiar with each other. The fact of familiarity, when clear, is noted in the annotations. Similarly, the annotations include information on whether the participant has a thorough (works or studies inside the building) or passing (works or studies in a nearby building) familiarity with the experiment site. 

In both sites, the receptionist would occupy the actual receptionist area in the lobby of the building. In Doha, on-duty security guards were present in the vicinity of the reception desk.

Each pair of participants would have 2-3 interactions with one of the subjects as a receptionist, and then they would switch roles and have 2 or 3 more interactions, depending on allotted time. After that, the participants were debriefed on their experiences. Overall, more than 60 interactions were recorded.

The interactions were recorded with 2 or 3 consumer-level high definition cameras. Visitor and receptionist were each dedicated a camera capturing their torso, arms and face that was positioned about 45 degrees off their default line of sight (namely, the line of sight that is  perpendicular to the front edge of the rectangular reception desk). Most of the interactions would have a third camera capturing the side view of the scene. All cameras were in plain view. In addition to the audio captured by the cameras, an audio recorder (iPod) was placed on the receptionist desk.

\section{Annotation scheme}

The main goal of our corpus is to analyze occurrences and timing of verbal and nonverbal behaviors. Consequently,  we have chosen to annotate the data at the level of granularity that minimizes the coding effort while at the same time allowing to capture timing and major features of communicative events. For example, instead of annotating each of preparation, hold, stroke, and retraction phases of a hand gesture~\cite{Kita1998} we annotate an interval between beginnings of the stroke and retraction phases. Similarly, facial expression are annotated as intervals approximately from the beginning of rise to the beginning of decay \cite{Hoque2011} phases, with some error inherent to manual annotation. The annotation scheme, developed in the process of annotating the corpus, is summarized in Table~\ref{table:annotation}.

\begin{table}[htb]
\begin{center}
\begin{tabular}{|l|p{2in}|}
\hline
Modality & Values \\ \hline\hline
Speech & Transcribed utterances, including non-words \\ \hline
Eye gaze & Pointing (self-initiated), pointing (following interlocutor), focus (interlocutor, guard, desktop, down, up, left, right, front, back, scattered, destination)\\ \hline
Face & smile (open or closed mouth)\\ \hline
Head & nod, half nod, double nod, multiple nod, upward nod, multiple upward nod, micro nod, shake \\ \hline
Hand & Pointing (left or right hand), finger only \\ \hline
Torso & Sitting, standing, focus (left, right, front, back, destination, interlocutor, desk) \\ \hline
\end{tabular}
\end{center}
\caption{Annotation scheme}
\label{table:annotation}
\end{table}

Coding nonverbal expressions, as well as transcribing ambiguous speech involves a degree of subjectivity. For example, the exact point of gaze fixation within the recipient's face is hard to identify even by the recipient himself~\cite{Cranach1973}. In fact, a typical direct eye contact consists of a sequence of fixations on different points on the face~\cite{Cook1977}. Since it is unclear whether the exact fixation pattern has any influence on social communication, in this study we do not distinguish between different fixation points within the general face area (neither does the video fidelity allow that). We plan to validate the annotations by employing a second annotator. 

The annotation is done using the multi-track video annotation tool Advene~\cite{Aubert2005}.

\section{Discussion}

\begin{figure*}[tb]
\centering
\includegraphics[width=6in]{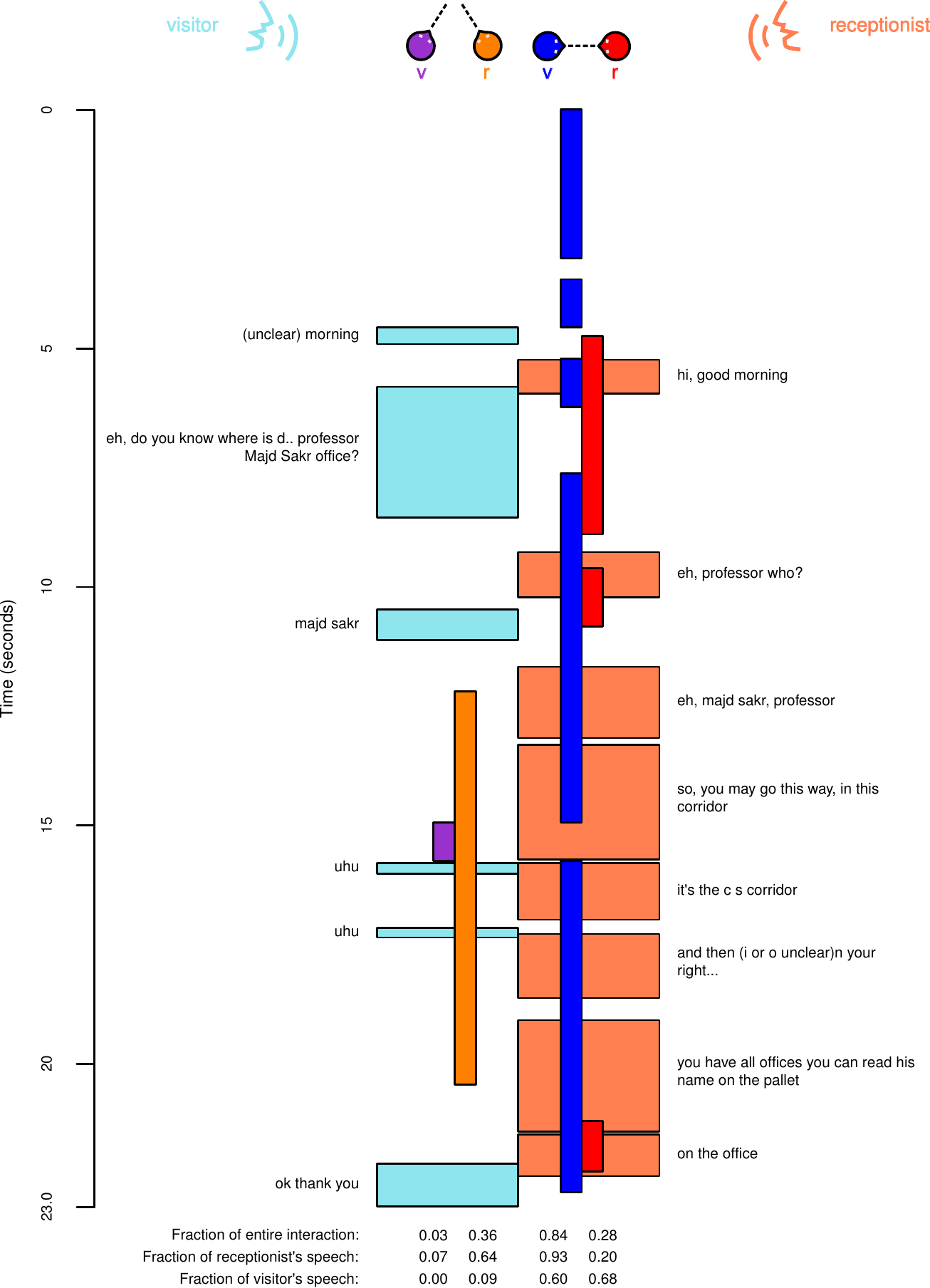}
\caption{Interaction between S1 as a visitor and S4 as a receptionist. Wide vertical stripes represent intervals of speech. Narrow vertical stripes represent (from left to right): intervals of visitor's and receptionist's gaze towards the direction pointed by the receptionist, and visitor's and receptionist's gaze towards each other. Color coding of these modalities is specified by the icons in the upper part of the plots.}
\label{fig:v1r4i3}
\end{figure*}

\begin{figure*}[tb]
\centering
\includegraphics[width=6in]{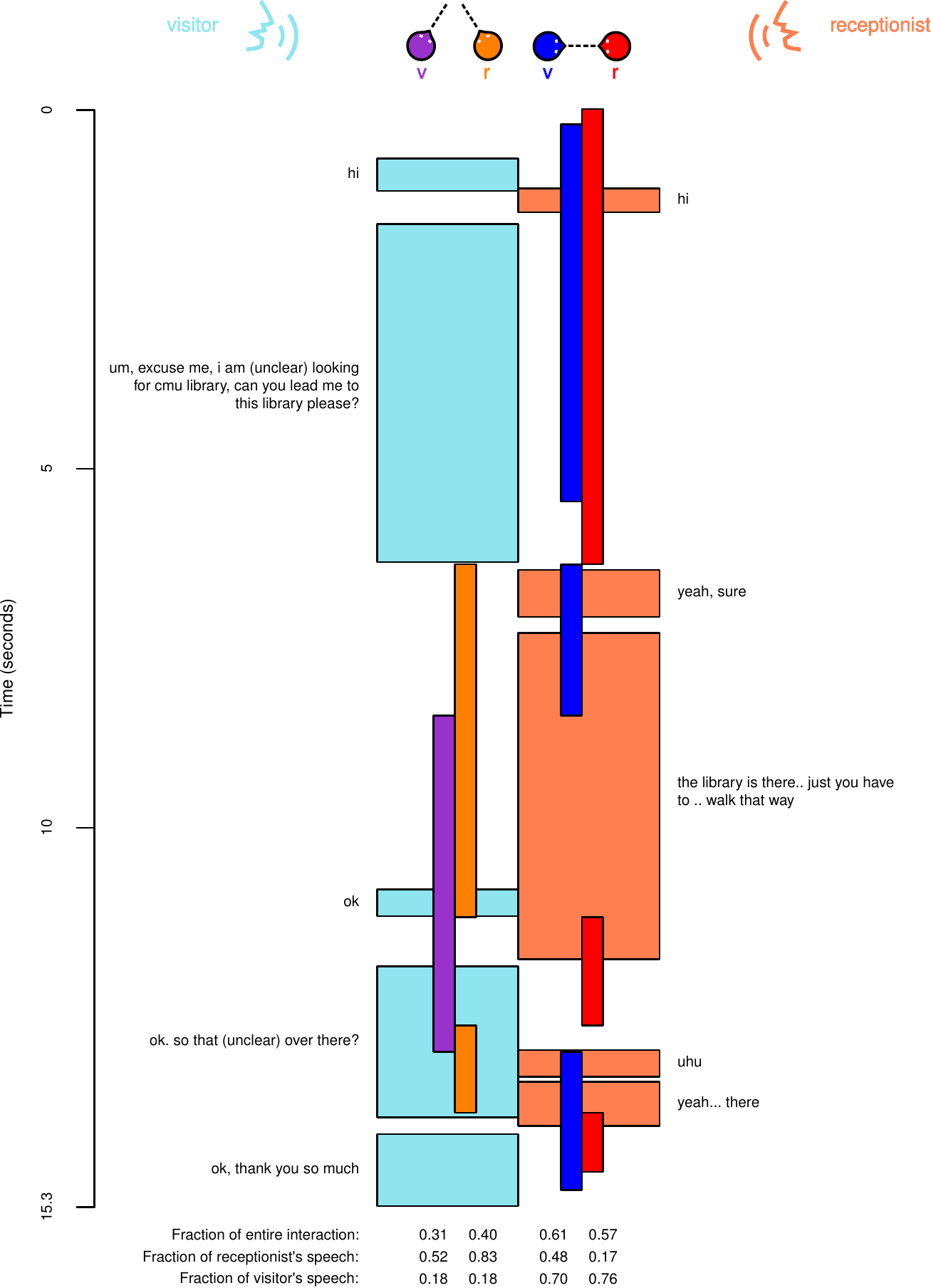}
\caption{Interaction between S4 as a visitor and S1 as a receptionist. 
Wide vertical stripes represent intervals of speech. Narrow vertical stripes represent (from left to right): intervals of visitor's and receptionist's gaze towards the direction pointed by the receptionist, and visitor's and receptionist's gaze towards each other.
Color coding of these modalities is specified by the icons in the upper part of the plots.}
\label{fig:v4r1i1}
\end{figure*}

While the small number of individual participants makes this corpus unsuitable for cross-subject analysis, the multiple trials may be accounted for by mixed-effects models~\cite{PinheiroBates2000}. More appropriately, the corpus should be used for qualitative analysis and formation of hypothesis for further studies. For example, compare the gaze behaviors of a native Arabic-speaking female S4 (Subject 4) playing a receptionist responding to native Arabic-speaking male S1 playing a visitor (Fig.~\ref{fig:v1r4i3}) versus the dialogue with the subjects' roles reversed (Fig.~\ref{fig:v4r1i1}). Notice that both subjects gazed at their interlocutor more in the visitor role. This appears to be a trend that can be explained in part by the receptionist looking towards the destination during the direction-giving speech, while the visitor may continue looking at the receptionist.

\begin{figure*}[tb]
\centering
\includegraphics[width=6in]{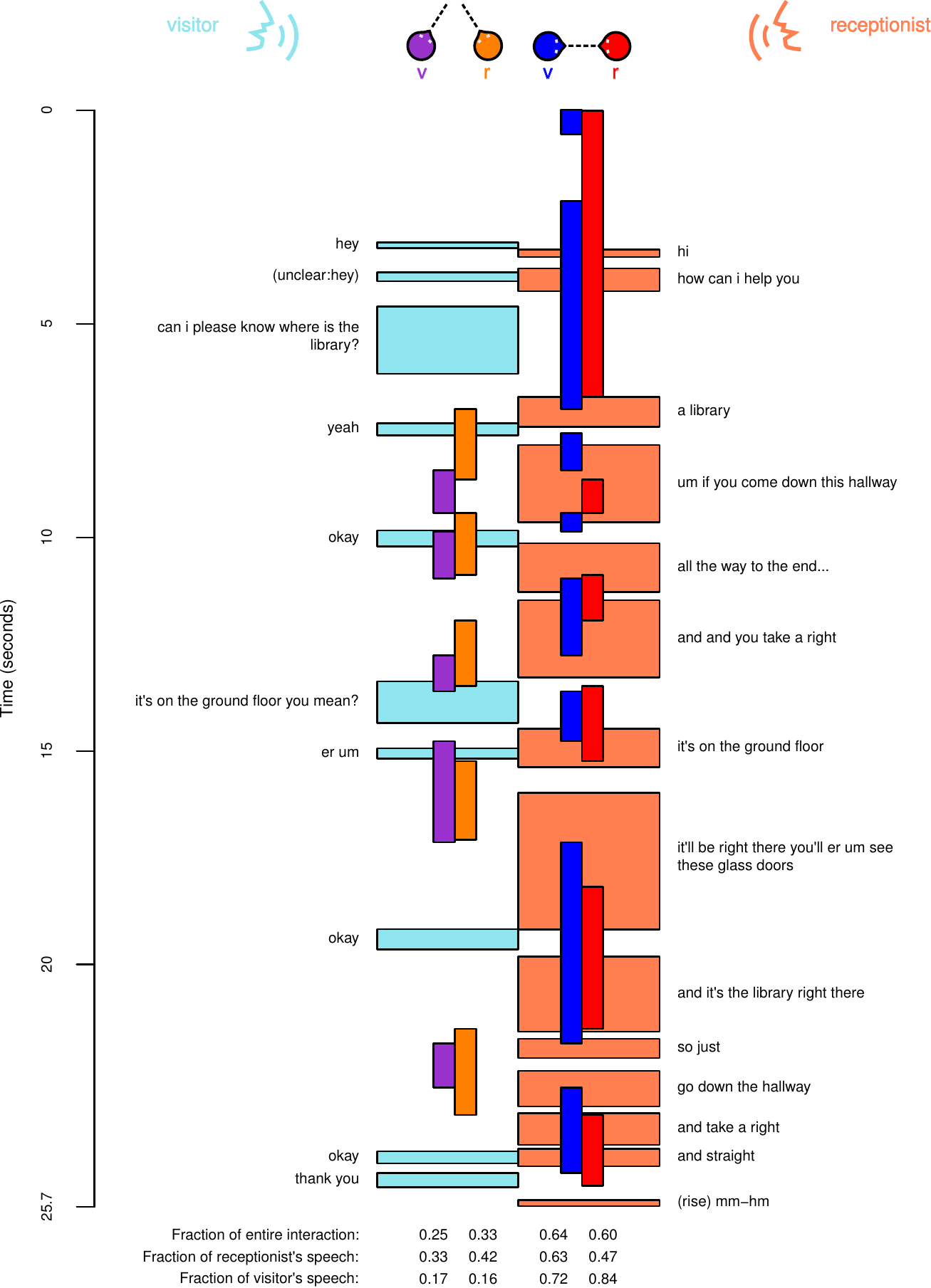}
\caption{Interaction between S11 as a visitor and S12 as a receptionist. The visitor's eye gaze for this particular dialogue is partially inferred from his head gaze. 
Wide vertical stripes represent intervals of speech. Narrow vertical stripes represent (from left to right): intervals of visitor's and receptionist's gaze towards the direction pointed by the receptionist, and visitor's and receptionist's gaze towards each other.
Color coding of these modalities is specified by the icons in the upper part of the plots.}
\label{fig:v11r12i3}
\end{figure*}

Now, compare a receptionist gaze of S4 (Fig.~\ref{fig:v1r4i3}) with one of S12 (Fig.~\ref{fig:v11r12i3}), who is a female native speaker of American English. Notice the short glances that punctuate fragments of the directions sequence spoken by S12. These glances appear to precede visitor's backchannels and therefore may play a role in connection events~\cite{Sidner2010}. Receptionist S4, on the contrary, did not glance at the visitor until the very end of the directions sequence. These different gaze behaviors may reflect individual styles, genders and cultures of receptionist-visitor pairs, or levels of comfort and expertise, among other possibilities. Further, more controlled, studies may address these hypothesis.

\section{Acknowledgments}

This publication was made possible by the support of an NPRP grant from the Qatar National Research Fund. The authors would like to express their gratitude to Michael Agar, Mark Barker, Justine Cassell, Anwar El-Shamy, Ismet Hajdarovic, Alicia Holland, Carol Miller, Dudley Reynolds, Michele de la Reza, Candace Sidner, Mark Stehlik, Mark C. Thompson, security and receptionist staff of CMU Qatar, and the study participants.


\bibliographystyle{abbrv}
\bibliography{../nl}

\end{document}